%% file: main.tex
\documentclass{llncs}
\usepackage[utf8]{inputenc}
\usepackage{amsmath}
\usepackage{amsfonts}
\usepackage{amssymb}

\usepackage{caption}
\usepackage{subcaption}
\usepackage{graphicx}
\usepackage[style=numeric-comp,
sorting=nyt,
doi=false,
url=false,
isbn=false,
firstinits,
maxbibnames=99,
backend=biber]{biblatex}
\bibliography{refs}

\title{Inducing Language Networks from Continuous Space Word Representations}
\author{Bryan Perozzi, Rami Al-Rfou, Vivek Kulkarni, Steven Skiena}
\institute{Department of Computer Science\\
Stony Brook University\\
 \{\email{bperozzi,ralrfou,vvkulkarni,skiena}\}\email{@cs.stonybrook.edu}}

\begin{document}
\maketitle

\abstract{Recent advancements in unsupervised feature learning have developed powerful latent representations of words.  However, it is still not clear what makes one representation better than another and how we can learn the ideal representation. Understanding the structure of latent spaces attained is key to any future advancement in unsupervised learning. 

In this work, we introduce a new view of continuous space word representations as language networks. We explore two techniques to create language networks from learned features by inducing them for two popular word representation methods and examining the properties of their resulting networks.  We find that the induced networks differ from other methods of creating language networks, and that they contain meaningful community structure.}

\input{00_introduction}

\input{01_neural_language_models}

\input{02_embedding_networks}

\input{03_discussion}

\input{04_related_work}

\input{05_conclustions}

\input{06_appendix}

\printbibliography

\end{document}

%% file: 00_introduction.tex
\section{Introduction}
Unsupervised feature learning (\emph{deep learning}) utilizes huge amounts of raw data to learn representations that model knowledge structure and disentangle the explanatory factors behind 
observed events.
Under this framework, symbolic sparse data is represented by lower-dimensional continuous spaces.
Integrating knowledge in this format is the secret behind many recent breakthroughs in machine learning based applications such as speech recognition, computer vision, and natural language processing (NLP) \cite{bengio2013representation}.

We focus here on word representations (\emph{word embeddings}) where each word representation consists of a dense, real-valued vector. During the pre-training stage, the representations acquire the desirable property that similar words have lower distance to each other than to unrelated words \cite{hinton2006reducing}.
This allows the representations to utilize the abundance of raw text available to learn features and knowledge that is essential for supervised learning applications such as part-of-speech tagging, named entity recognition, machine translation, language modeling, sentiment analysis etc \cite{collobert2011natural,schwenk2012large,mikolov2011extensions,GlorotICML2011}.

Several methods and algorithms have been proposed to learn word representations along different benchmarks for evaluation \cite{expressive}.
However, these evaluations are hard to comprehend as they squash the analysis of the representation's quality into abstract numbers.
To enable better understanding of the actual structure of word relationships which have been captured, we have to address the problems that come with analyzing high-dimensional spaces (typically between 50-1000 dimensions).
We believe that network induction and graph analysis are appropriate tools to give us new insights.

In this work, we seek to induce meaningful graphs from these continuous space language models. Specifically, our contributions include:
\begin{itemize}
\item \textbf{Analysis of Language Network Induction} - We propose two criteria to induce networks out of continuous embeddings. For both methods, we study and analyze the characteristics of the induced networks. Moreover, the networks generated lead to easy to understand visualizations.
\item \textbf{Comparison Between Word Representation Methods} - We evaluate the quality of two well known words embeddings. We contrast between their characteristics using the analysis developed earlier.
\end{itemize}

The remainder of this paper is set up as follows.  First, in Section \ref{section.nml}, we describe continuous space language models that we consider.
In Section \ref{section.network}, we discuss the choices involved with inducing a network from these embeddings and examine the resulting networks.
Finally, we finish with a discussion of future work and our conclusions.

%% file: 01_neural_language_models.tex
\section{Continuous Space Language Models}
\label{section.nml}

The goal of a language model is to assign a probability for any given sequence of words estimating the likelihood of observing such a sequence.
The training objective usually maximizes the joint probability of the training corpus.
A continuous space probabilistic language model aims to estimate such probability distribution by, first, learning continuous representations for the words and phrases observed in the language.
Such mapping is useful to cope with the curse of dimensionality in cases where data distribution is sparse as natural language.
Moreover, these representations could be used as features for natural language processing applications, domain adaptation and learning transfer scenarios that involve text or speech.

More precisely, given a sequence of words $S = [w_1 \dots w_{k}]$, we want to maximize $P(w_1, \dots, w_k)$ and learn representations for words.
During the training process the continuous space language model learns a mapping of words to points in $\mathbb{R}^d$, where $d$ usually ranges between $20-200$.
Prior to training we build a vocabulary $V$ that consists of the most frequent $|V|$ words, we map each word to a unique identifier that indexes an embeddings matrix $C$ that has a size of $|V| \times d$.
The sequence $S$ is now represented by a matrix $\begin{bmatrix}C[w_1]^T& \dots & C[w_k]^T\end{bmatrix}^T$, enabling us to compose a new representation of the sequence using one of several compositional functions.
The simplest is to concatenate all the rows in a bigger vector with size $kd$.
Another option is to sum the matrix row-wise to produce a smaller representation of size $d$. While the first respects the order of the words, it is more expensive to compute.

Given a specific sequence representation as an input, we will define a task that the model should solve, given the sequence representation as the only input.
Our choice of the task ranges from predicting the next/previous word(s) to distinguishing between observed phrases and other corrupted copies of them.
The chosen task and/or the compositional function influence the learned representations greatly as we will discuss later.

We will focus our investigations, here, on two embeddings which are trained with different tasks and compositional functions; the Polyglot and SkipGram embeddings.

\subsection{Polyglot}
The Polyglot project offers word representations for each language in Wikipedia \cite{polyglot}.  For large enough Wikipedias, the vocabulary consists of the most frequent 100,000 words.
The representations are learned through a procedure similar to the one proposed by \textcite{collobert2011natural}. For a given sequence of words $S_t = [w_{t-k} \dots w_t \dots w_{t+k}]$ observed in the corpus $T$, a corrupted sequence $S_t^{\prime}$ will be constructed by replacing the word in the middle $w_t$ with a word $w_j$ chosen randomly from the vocabulary $V$. Once the vectors are retrieved, we compose the sequence representation by concatenating the vectors into one vector called the projection layer $S_t$. The model is penalized through the hinge loss function,

\begin{equation*}
\frac{1}{T} \sum_{t=1}^{t=T} | 1 - score(S_t^{\prime}) + score(S_t)|_+
\end{equation*}
where $score$ is calculated through a hidden layer neural network
\begin{equation*}
score(S_t) = W_2 (tanh(W_1 S_t + b_1)) + b_2.
\end{equation*}

For this work, we use the Polyglot English embeddings\footnote{Polyglot embeddings and corpus available at \texttt{\url{http://bit.ly/embeddings}}} which consist of the 100,000 most frequent words in the English Wikipedia, each represented by a vector in $\mathbb{R}^{64}$. 

\subsection{SkipGram}
While the Polyglot embeddings consider the order of words to build the representation of any sequence of words, the SkipGram model proposed by \textcite{mikolov2013efficient} maximizes the average log probability of the context words independent of their order

\begin{equation*}
\frac{1}{T} \sum_{t=1}^{T} \Big[\sum_{j=-k}^{k} \log p(w_{t+j}|w_t)\Big]
\end{equation*}
where $k$ is the size of the training window. This allows the model to scale to larger context windows. 
In our case, we train a SkipGram model\footnote{SkipGram training tool available at \texttt{\url{https://code.google.com/p/word2vec/}}} on the English Wikipedia corpus offered by the Polyglot project for the most frequent 350,000 words with context size $k$ set to 5 and the embeddings vector size set to 64.


\subsection{Random}
In order to have a baseline, we also generate random embeddings for the most frequent 100,000 words.  The initial position of words in the Polyglot embeddings were sampled from a uniform distribution, therefore, we generate the random embedding vectors by sampling from $\mathcal{U}(\bar{m} - \sigma, \bar{m} + \sigma)$, where $\bar{m}$ and $\sigma$ are the mean and standard deviation of the trained Polyglot embeddings' values respectively.  This baseline allows us to see how the language networks we construct differ from networks induced from randomly initialized points.

%% file: 02_embedding_networks.tex
\section{Word Embedding Networks}
\label{section.network}

We now consider the problem of constructing a meaningful network given a continuous space language model.  As there are a variety of ways in which such a network could be induced, we start by developing a list of desirable properties for a language network.  Specifically, we are seeking to build a network which:
\begin{enumerate}
\item{\textbf{Is Connected} - In a connected graph, all the words can be related to each other. This allows for a consistent approach when trying to use the network to solve real-world problems.}
\item{\textbf{Has Low Noise} - Minimizing the spurious correlations captured by our discrete representation will make it more useful for application tasks.}
\item{\textbf{Has Understandable Clusters} - We desire that the community structure in the network reflects the syntactic and semantic information encoded in the word embeddings.}
\end{enumerate}

We also require a method to compute the distance in the embedding space.  While there are a variety of metrics that could be used, we found that Euclidean distance worked well.  So we use:

\begin{equation}
dist(x,y) = ||x-y||_2^2 = (\sum^{m}_{i=1}(x_i - y_i)^2)^{(1/2)}
\end{equation}

where $x$ and $y$ are words in an $d$-dimensional embedding space ($x,y \in \mathbb{R}^d$).  With these criteria and a distance function in hand, we are ready to proceed.  We examine two approaches for constructing graphs from word embeddings, both of which seek to link words together which are close in the embedding space.  For each method, we induce networks for the $20,000$ most frequent words for each embedding type, and compare their properties.

\subsection{$k$-Nearest Neighbors}

The first approach we will consider is to link each word to the $k$ closest points in the embedding space.  More formally, we induce a set of directed edges through this method:

\begin{equation}
E_{knn} = \{(u,v) : \min_x dist(u,v) \} \;\;\; \forall u,v \in V, x \leq k
\end{equation}

where $\min_x$ denotes the rank of the $x$-th number in ascending sorted order (e.g. $\min_0$ is the minimum element, $\min_1$ the next smallest number).  After obtaining a directed graph in this fashion, we convert it to an undirected one.

\begin{figure}[htb!]
        \centering
        \begin{subfigure}[b]{0.32\textwidth}
                \includegraphics[width=\textwidth]{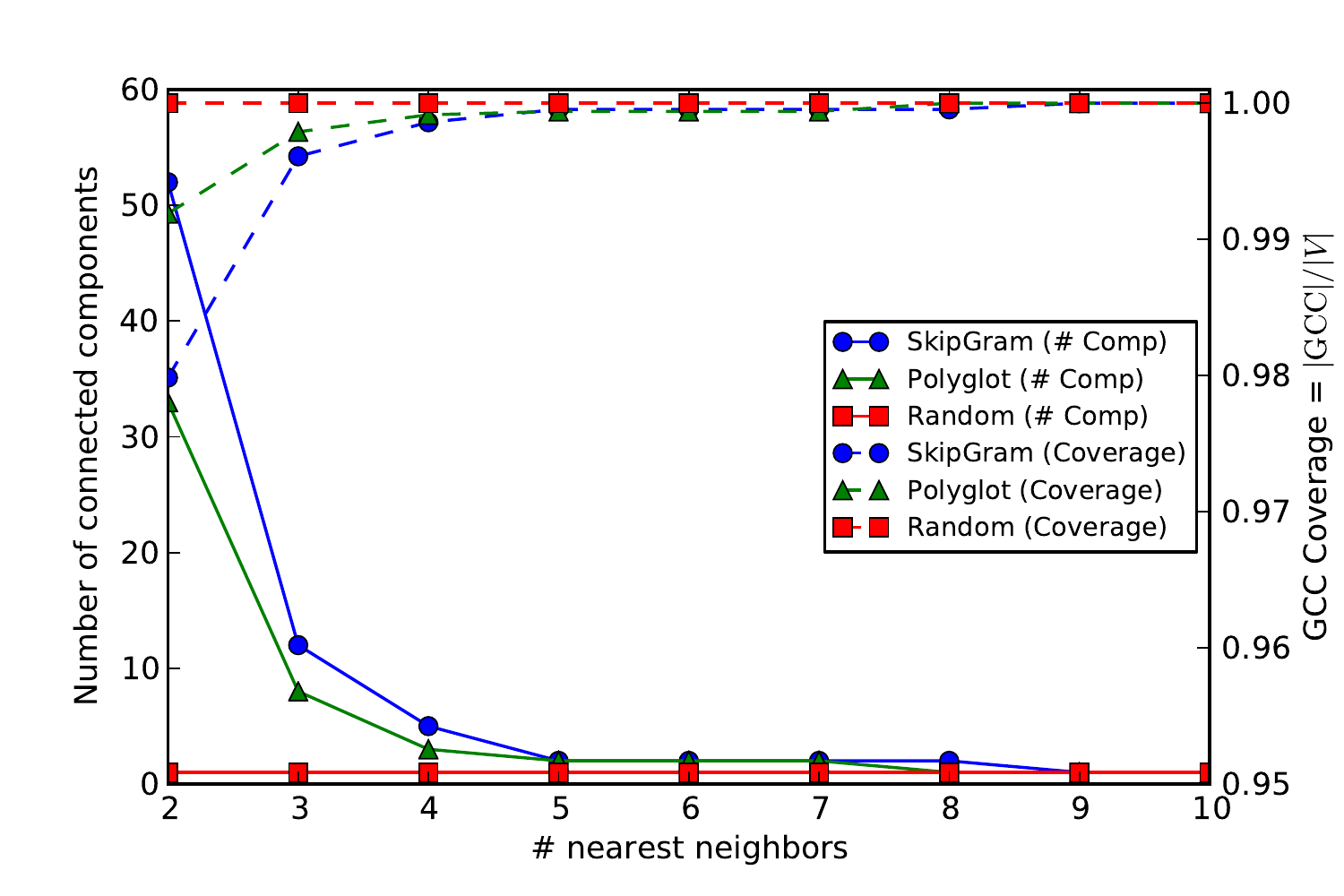}
                \caption{$k$-NN Coverage}
                \label{fig:knn_coverage}
        \end{subfigure}        
        \begin{subfigure}[b]{0.32\textwidth}
                \includegraphics[width=\textwidth]{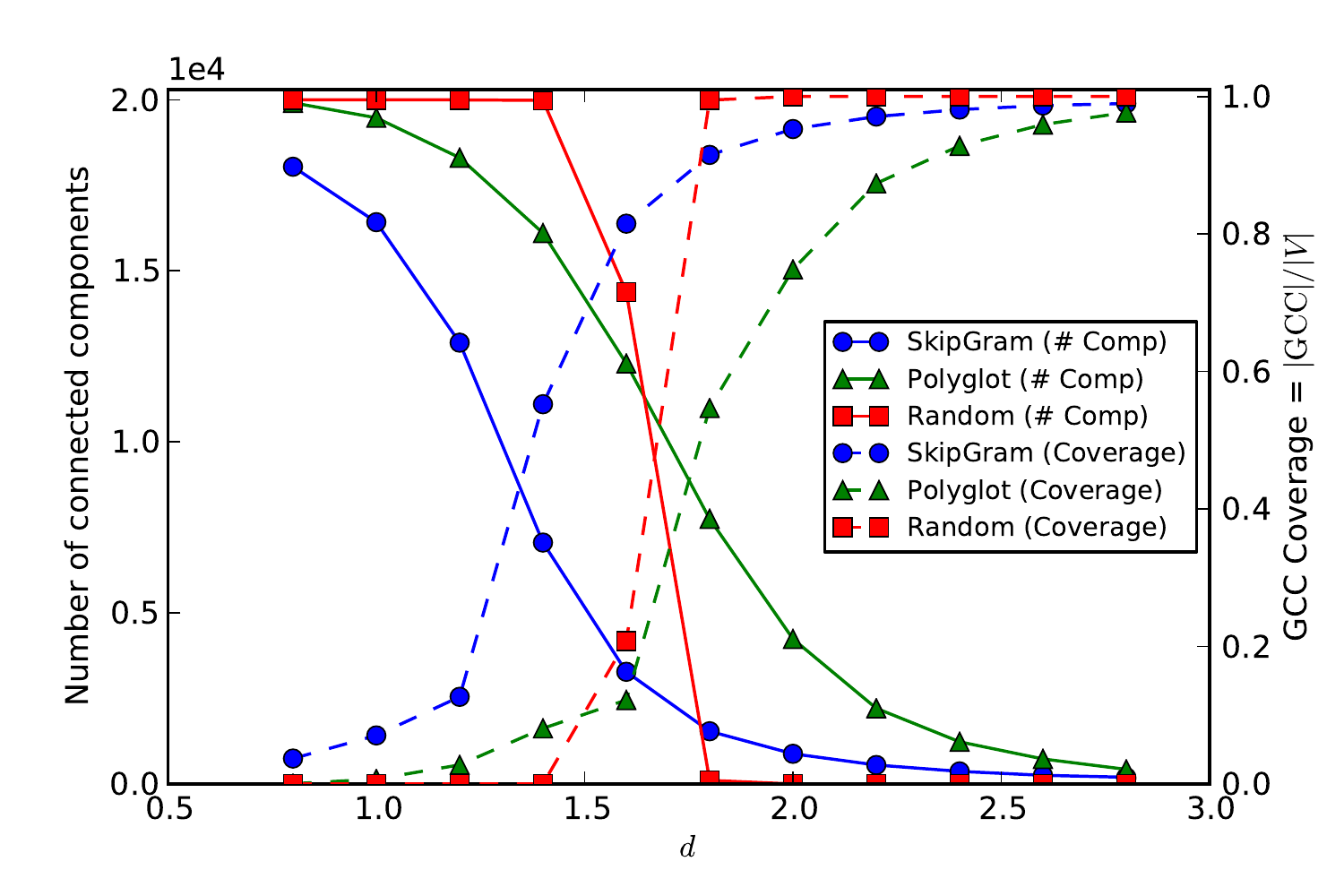}
                \caption{$d$-Threshold Coverage}
                \label{fig:d_coverage}
        \end{subfigure}
	\caption{\textbf{Graph Coverage}. The connected components and relative size of the Giant Connected Component (GCC) in graphs created by both methods.  We see that very low values of $k$ quickly connect the entire network (\ref{fig:knn_coverage}), while values of $d$ appear to have a transition point before a GCC emerges (\ref{fig:d_coverage}).}
	\label{fig:method_diff}	
\end{figure}

The resulting undirected graph does not have a constant degree distribution.  This is due to the fact that the nearest-neighbor relation may not be symmetric.  Although all vertices in the original directed graph have an out-degree of $k$, their orientation in the embedding space means that some vertices will have higher in-degrees than others.  

Results from our investigation of basic network properties of the $k$-NN embedding graphs are shown in Figures \ref{fig:method_diff} and \ref{fig:community_metrics}.  In (\ref{fig:knn_coverage}) we find that the embedding graphs have few disconnected components, even for small values of $k$.  In addition, there is an obvious GCC which quickly emerges. In this way, the embeddings are similar to the network induced on random points (which is fully connected at $k=2$).  
We performed an investigation of the smaller connected components when $k$ was small, and found them to contain dense groupings of words with very similar usage characteristics (including ordinal values, such as Roman numerals (\texttt{II,III,IV})).

In (\ref{fig:clustering_coeff}) we see that the clustering coefficient initially grows quickly as we add edges to our network ($k \leq 6$), but has leveled off by ($k=20$).  This tendency to bridge new clusters together, rather than just expand existing ones, may be related to the \emph{instability} of the nearest neighbor \cite{beyer1999nearest} in high dimensional spaces.  In (\ref{fig:knn_q}), we see that the networks induced by the $k$-NN are not only connected, but have a highly modular community structure.

\subsection{$d$-Proximity}

The second approach we will consider is to link each word to all those within a fixed distance $d$ of it:

\begin{equation}
E_{proximity} = \{(u,v) : dist(u,v) < d \} \;\;\; \forall u,v \in V
\end{equation}

\begin{figure}[htb!]
        \centering
        \begin{subfigure}[b]{0.32\textwidth}
                \includegraphics[width=\textwidth]{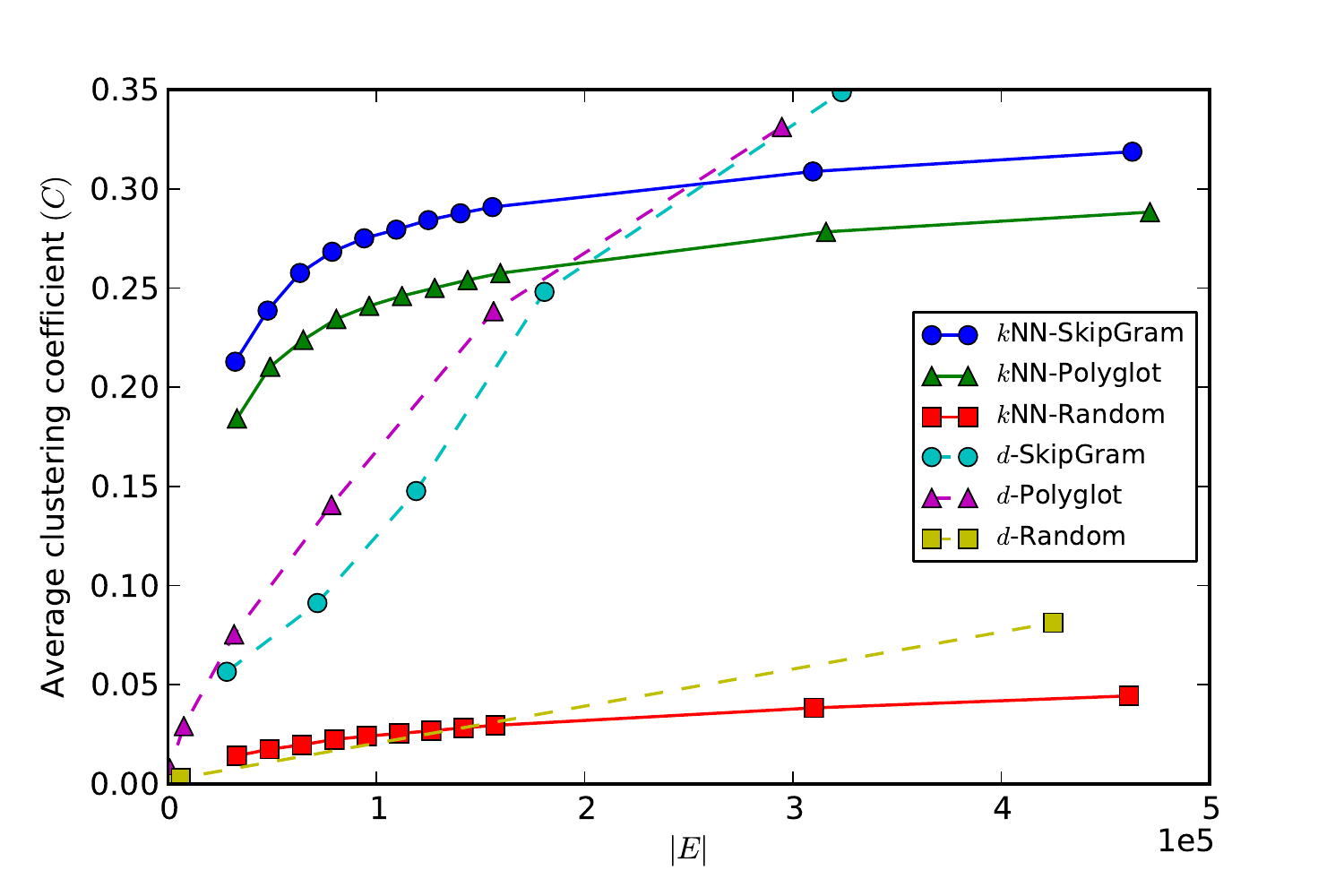}
                \caption{Clustering Coeff., $C$}
                \label{fig:clustering_coeff}
        \end{subfigure}        
        \begin{subfigure}[b]{0.32\textwidth}
                \includegraphics[width=\textwidth]{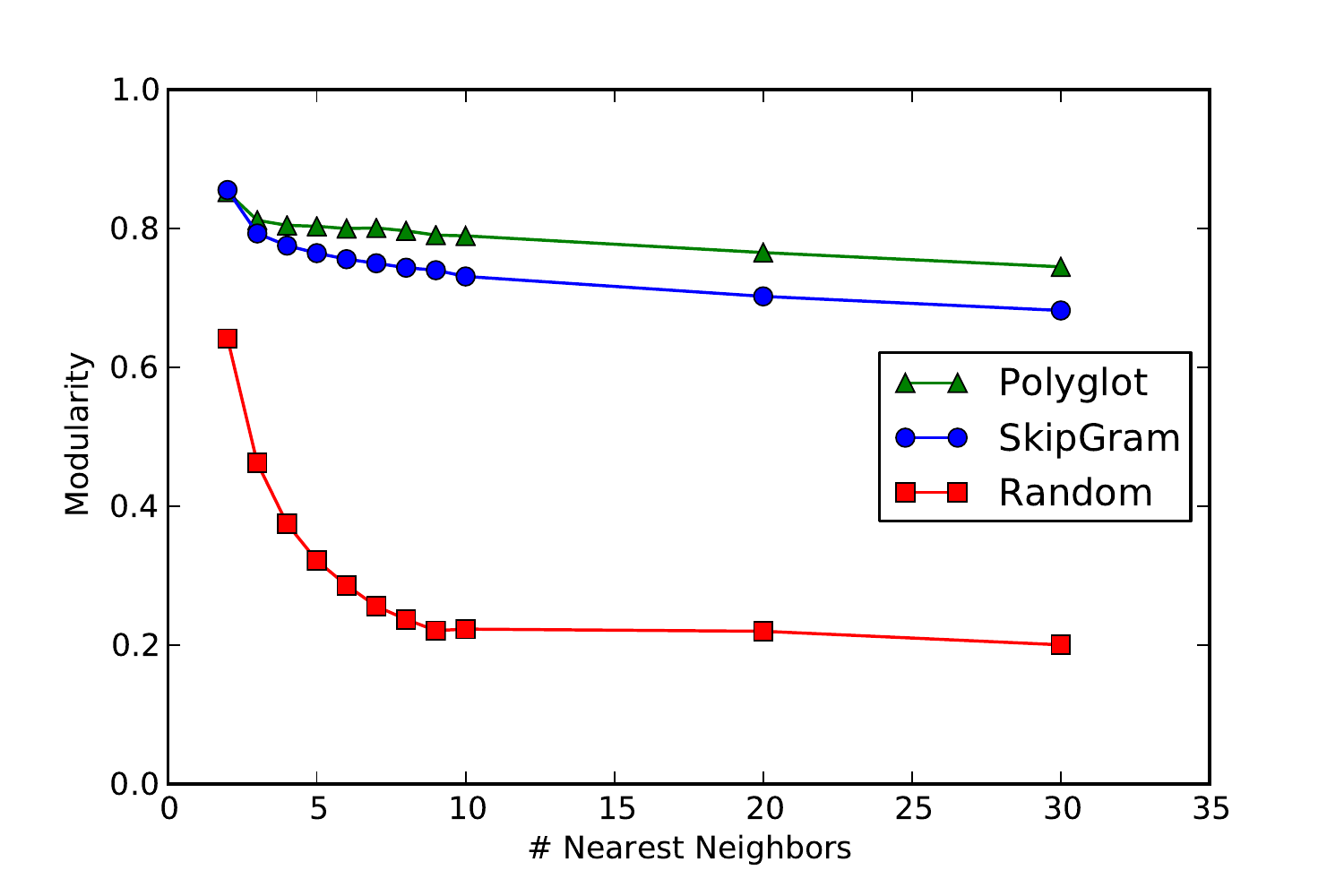}
                \caption{$k$-NN Modularity, $Q_{knn}$}
                \label{fig:knn_q}
        \end{subfigure}        
	\caption{\textbf{Community Metrics}. In (\ref{fig:clustering_coeff}), $C$ shown for $k$ = [2,30] and $d$ = [0.8,1.6] against number of edges in the induced graph.  When the total number of edges is low ($|E| < 150,000$), networks induced through the $k$-NN method have more closed triangles than those created through $d$-Proximity.  In (\ref{fig:knn_q}), $Q_{knn}$ starts high, but slowly drops as larger values of $k$ include more spurious edges. }
	\label{fig:community_metrics}
\end{figure}

We perform a similar investigation of the network properties of embedding graphs constructed with the $d$-Proximity method.  The results are shown in Figures \ref{fig:method_diff} and \ref{fig:community_metrics}.  We find that networks induced through this method quickly connect words that are near each other in the embedding space, but do not bridge distant groups together.  They have a large number of connected components, and connecting 90\% of the vertices requires using a relatively large value of $d$ (\ref{fig:d_coverage}).  

The number of connected components is closely related to the average distance between points in the embedding space (around $d=$(3.25, 3.80, 2.28) for (SkipGram, Polyglot, Random)).  As the value of $d$ grows closer to this average distance, the graph quickly approaches the complete graph.  

Figure \ref{fig:clustering_coeff} shows that as we add more edges to the network, we add triangles at a fast rate than using the $k$-NN method.

\subsection{Discussion}
Here we discuss the differences exposed between the methods for inducing word embeddings, and the differences exposed between the embeddings themselves.

\subsubsection{Comparison of Network Induction Methods.}
\begin{figure}[htb!]
	\includegraphics[width=\textwidth,keepaspectratio=true]{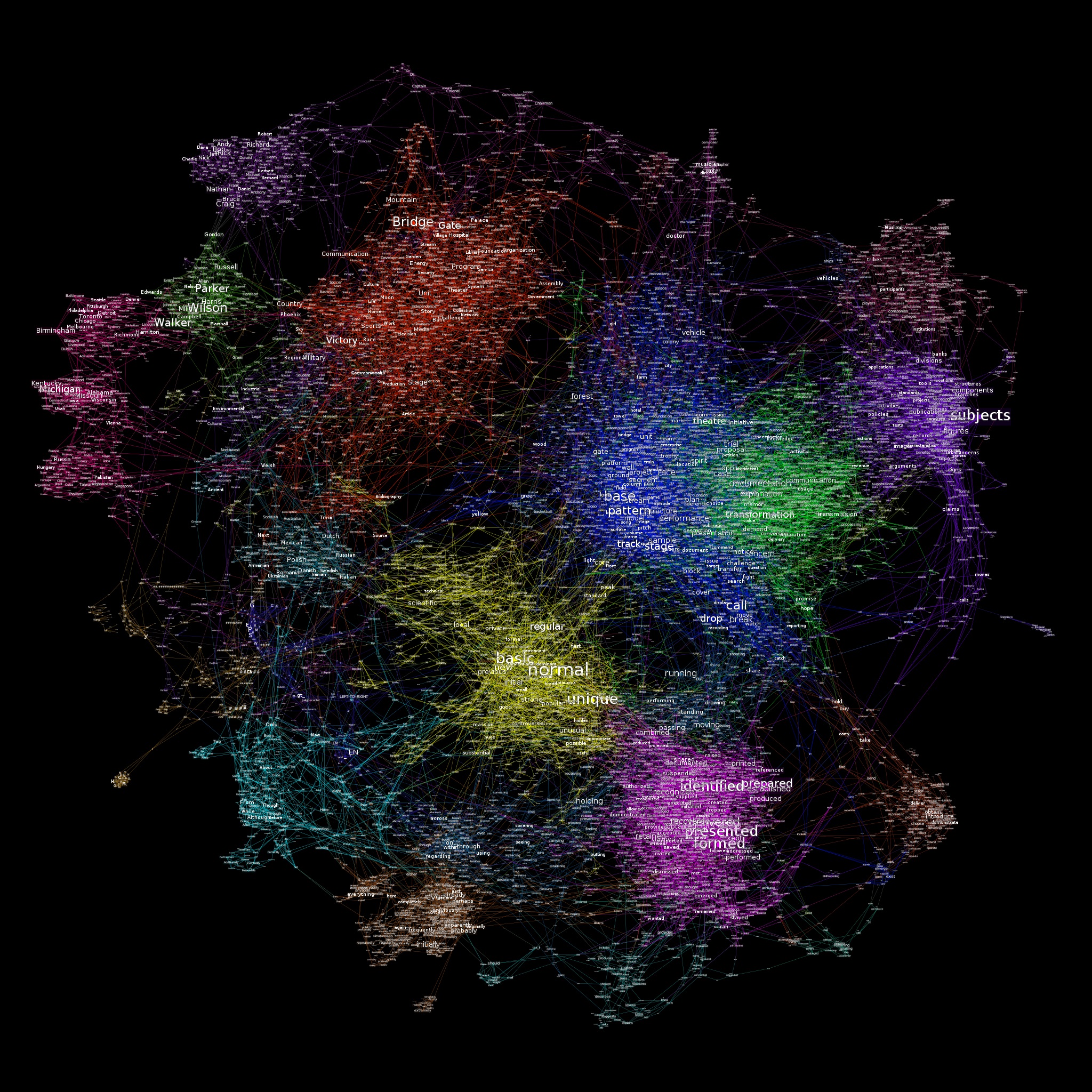}
	\caption{\textbf{Polyglot Nearest Neighbor Graph}. Here we connect the nearest neighbors ($k=6$) of the top 5,000 most frequent words from the Polyglot English embeddings. Shown is the giant connected component of the resulting graph ($|V| = 11,239$; $|E| = 26,166$). Colors represent clusters found through the Louvain method (modularity $Q = 0.849$).  Vertex label size is determined by its PageRank.  Best viewed in color.}        
	\label{fig:polyglot_en}
\end{figure}

Which method then, provides the better networks from word embeddings?  To answer this question, we will use the properties raised at the beginning of this section:

\begin{enumerate}
\item{\textbf{Connectedness} - Networks induced through the $k$-NN method connect much faster (as a function of edges) than those induced through $d$-Proximity (Fig. \ref{fig:method_diff}).  Specifically, the network induced for $k=6$ has nearly full coverage (\ref{fig:knn_coverage}) with only 100K edges (\ref{fig:clustering_coeff}).}
\item{\textbf{Spurious Edges} - We desire that our resulting networks should be modular.  As such we would prefer to add edges between members of a community, instead of bridging communities together.  For low values of $|E|$, the $k$-NN approach creates networks which have more closed triangles (\ref{fig:clustering_coeff}).  However this does not hold in networks with more edges.}
\item{\textbf{Understandable Clusters} - In order to qualitatively examine the quality of such a language network, we induced a subgraph with the $k$-NN of the most frequent 5,000 words in the Polyglot embeddings for English. 
Figure \ref{fig:polyglot_en} presents the language network constructed for ($k=6$).  }
\end{enumerate}

According to our three criteria, $k$-NN seems better than $d$-Proximity.  In addition to the reasons we already listed, we prefer $k$-NN as it seems to require less parameterization ($d$-Proximity has a different optimal $d$ for each embedding type).

\subsubsection{Comparison of Polyglot and SkipGram.}
Having chosen to use $k$-NN as our preferred method for inducing language networks, we now examine the difference between the Polyglot and SkipGram networks.  

\emph{Clustering Coefficient.}
We note that in Figure \ref{fig:clustering_coeff}, the SkipGram model has a consistently higher clustering coefficient than Polyglot in $k$-NN networks.  A larger clustering coefficient denotes more triangles, and this may indicate that points in the SkipGram space form more cohesive local clusters than those in Polyglot.  Tighter local clustering may explain some of the interesting regularities observed in the SkipGram embedding \cite{mikolov2013linguistic}.

\emph{Modularity.}  In Figure \ref{fig:knn_q}, we see that Polyglot modularity is consistently above the SkipGram modularity.  SkipGram's embeddings capture more semantic information about the relations between words, and it may be that causes a less optimal community structure than Polygot whose embeddings are syntactically clustered.

\emph{Clustering Visulizations.} In order to understand the differences between the language networks better, we conducted an examination of the clusters found using the Louvain method \parencite{blondel2008fast} for modularity maximization. Figure \ref{fig:polyglot_en_closeup} examines communities from both Polyglot and SkipGram in detail.

%% file: 03_discussion.tex

\begin{figure}
        \centering
        \begin{subfigure}[b]{0.48\textwidth}
                \includegraphics[width=\textwidth]{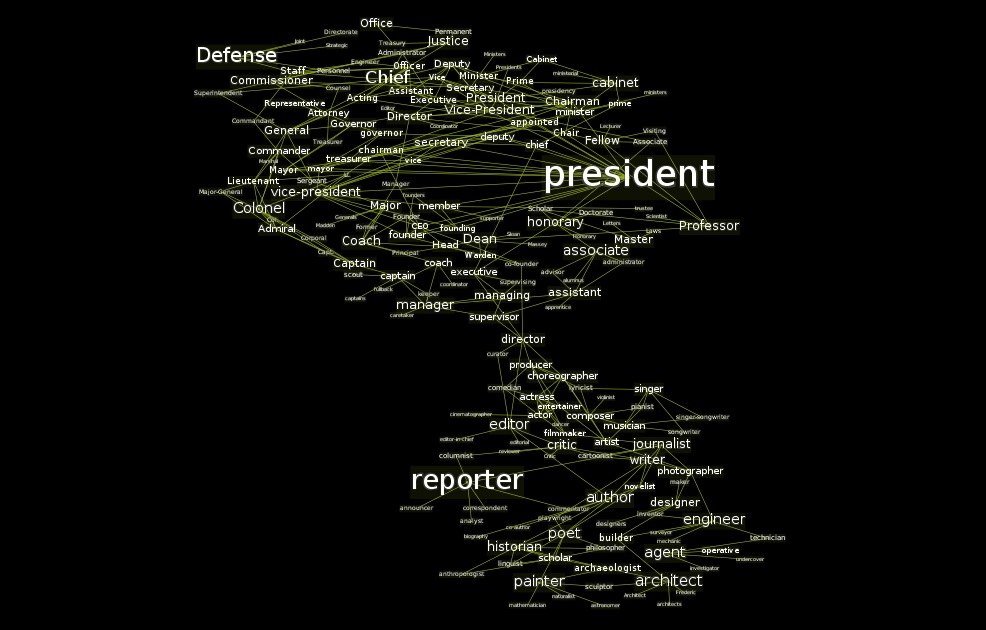}
                \caption{Professions (SkipGram)}
                \label{fig:tiger}
        \end{subfigure}
        ~
        \begin{subfigure}[b]{0.48\textwidth}
                \includegraphics[width=\textwidth]{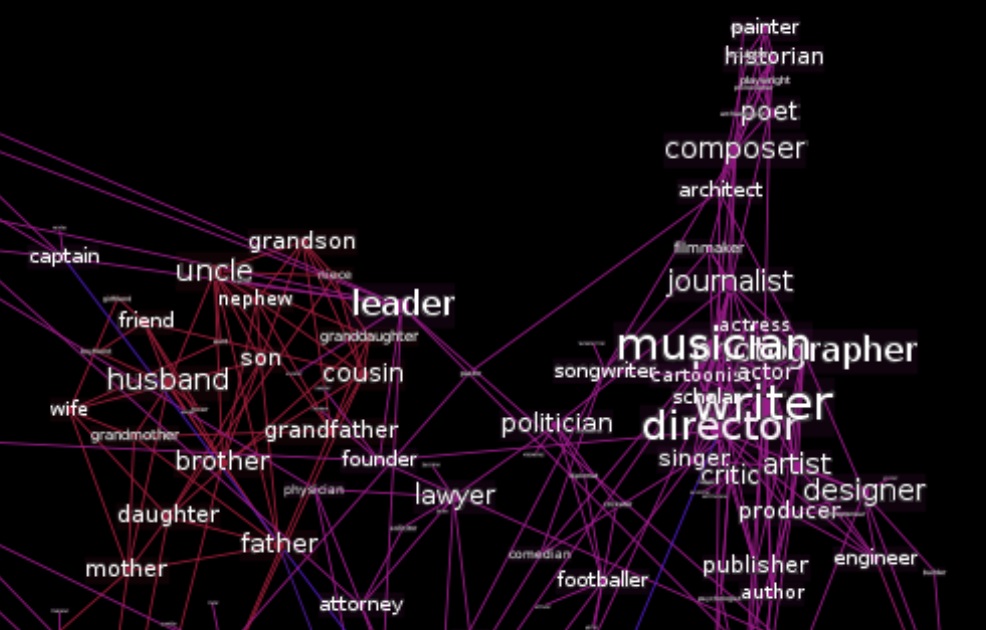}
                \caption{Professions (Polyglot)}
                \label{fig:tiger}
        \end{subfigure}                
        ~
        \begin{subfigure}[b]{0.48\textwidth}
                \includegraphics[width=\textwidth,trim=0 15 0 0,clip=true]{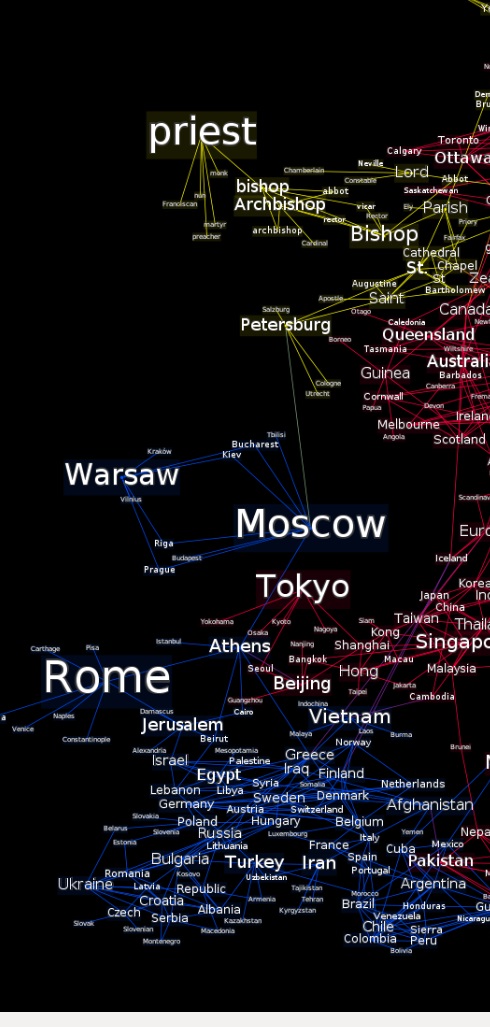}
                \caption{Locations (SkipGram)}
                \label{fig:sg_loc}
        \end{subfigure}
        ~
        \begin{subfigure}[b]{0.48\textwidth}
                \includegraphics[width=\textwidth]{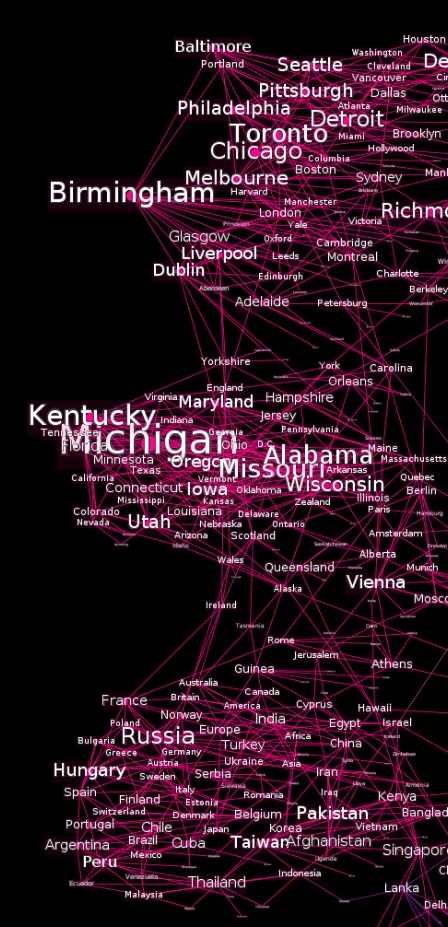}
                \caption{Locations (Polyglot)}
                \label{fig:mouse}
        \end{subfigure}        
        \caption{Comparison of clusters found in Polyglot and SkipGram language networks.  Polyglot clusters viewed in context of the surrounding graph, SkipGram clusters have been isolated to aide in visualization.   SkipGram's bag-of-words approach favors a more semantic meaning between words, which can make its clusters less understandable  (Note how in Figure \ref{fig:sg_loc} \texttt{Petersburg} is included in a cluster of religious words, because of \texttt{Saint}.)   Images created with Gephi \parencite{ICWSM09154}.}\label{fig:polyglot_en_closeup}
\vspace{-0.5cm}	
\end{figure}

%% file: 04_related_work.tex
\section{Related Work}

Here we discuss the relevant work in language networks, and word embeddings.  There is also related work on the theoretical properties of nearest neighbor graphs, consult \textcite{eppstein1997nearest} for some basic investigations.

\subsection{Language Networks}

\emph{Word Co-occurrences.} One branch of the study of language as networks seeks to build networks directly from a corpus of raw text.
\textcite{i2001small} examine word co-occurrence graphs as a method to analyze language.  In their graph, edges connect words which appear below a  fixed threshold ($d \leq 2$) from each other in sentences.  They find that networks constructed in this manner show both small world structure, and a power law degree distribution.  Language networks based on word co-occurrence have been used in a variety of natural language processing tasks, including motif analysis of semantics \parencite{biemann-roos-weihe:2012:PAPERS}, text summarization \parencite{Antiqueira2009584} and resolving disambiguation of word usages \parencite{Veronis2004223}.

\emph{Hypernym relations.}
Another approach to studying language networks relies on studying the relationships between words exposed by a written language reference.  \textcite{PhysRevE.65.065102} use a thesaurus to construct a network of synonyms, which they find to find to exhibit small world structure.  
In \parencite{sigman2002global}, \citeauthor{sigman2002global} investigate the graph structure of the Wordnet lexicon.  They find that the semantic edges in Wordnet follow scale invariant behavior and that the inclusion of polysemous edges drastically raises the clustering coefficient, creating a small world effect in the network.

\emph{Relation to our work.}
Much of the previous work in language networks build networks that are prone to noise from spurious correlations in word co-occurrence or infrequent word senses \parencite{i2001small,sigman2002global}.
Dimensionality reduction techniques have been successful in mitigating the effects of noise in a variety of domains.
The word embedding methods we examine are a form of dimensionality reduction that has improved performance on several NLP tasks and benchmarks.

The networks produced in our work are considerably different from language networks created by previous work that we are aware of.
We find that our degree distribution does appear to follow a power-law (like \parencite{i2001small,PhysRevE.65.065102,sigman2002global}) and we have some small world properties like those present in those works (such as $C \gg C_{random}$).
However, the average path length in our graphs is considerably larger than the average path length in random graphs with the same node and edge cardinalities.
Table \ref{table.comparisons} shows a comparison of metrics from different approaches to creating language networks.\footnote{Our induced networks available at \url{http://bit.ly/inducing_language_networks}} 

\begin{table}[htb]
  \centering
    \begin{tabular}{l c c c c c c c}
		\hline
		& $|V|$ & $|E|$ & $C$ & $C_{random}$ & $pl$ & $pl_{random}$& $\gamma$  \\
		\hline
    		\textcite{i2001small}(UWN) & $478,773$ & $1.77 \times 10^7$ & 0.687 & $1.55 \times 10^{-4}$ & $2.63^*$ & $3.03$& -1.50,-2.70 \\
    		\textcite{i2001small}(RWN) & $460,902$ & $1.61 \times 10^7$ & 0.437 & $1.55 \times 10^{-4}$ & $2.67^*$ & $3.06$& -1.50,-2.70\\
    		\textcite{PhysRevE.65.065102} & $30,244$ & $-$ & 0.53 & $0.002$ & $3.16$ & $-$ & $-$\\    		
    		\hline
    		\hline   		
    		Polyglot, $6$-NN & $20,000$ & $96,592$ & 0.241 & $0.0004$ & $6.78^*$ & $4.62^*$ & -1.31\\
    		SkipGram, $6$-NN & $20,000$ & $94,172$ & 0.275 & $0.0004$ & $6.57^*$ & $4.62^*$ & -1.32\\ 		
		\hline
		\\
    \end{tabular}
  \caption{A comparison of properties of language networks from the literature against those induced on the 20,000 most frequent words in the Polyglot and SkipGram Embeddings. ($C$ clustering coefficient, $pl$ average path length, $\gamma$ exponent of power law fits to the degree distribution) `*' denotes values which have been estimated on a random subset of the vertices.}
  \label{table.comparisons}
\end{table}

\subsection{Word Embeddings}

Distributed representations were first proposed by \textcite{hinton1986learning}, to learn a mapping of symbolic data to continuous space.
These representations are able to capture fine grain structures and regularities in the data  \cite{mikolov2013linguistic}.
However, training these models is slow due to their complexity.
Usually, these models are trained using back-propagation algorithm \cite{rumelhart1986learning} which requires large amount of computational resources.
With the recent advancement in hardware performance, \textcite{bengio2006neural} used the distributed representations to produce a state-of-the-art probabilistic language model.
The model maps each word in a predefined vocabulary $V$ to a point in $\mathbb{R}^d$ space (word embeddings). The model was trained on a cluster of machines for days. 
More applications followed, \textcite{collobert2011natural} developed SENNA, a system that offers part of speech tagger, chunker, named entity recognizer, semantic role labeler and discriminative syntactic parser using the distributed word representations.
To speed up the training procedure, importance sampling \cite{bengio2008adaptive} and hierarchical softmax models \cite{morin2005hierarchical,mnih2008scalable} were proposed to reduce the computational costs.
The training of word representations involves minimal amount of language specific knowledge and expertise.
\textcite{polyglot} trained word embeddings for more than a hundred languages and showed that the representations help building multilingual applications with minimal human effort.
Recently, SkipGram and Continuous bag of words models were proposed by \textcite{mikolov2013efficient} as simpler and faster alternatives to neural network based models. 

%% file: 05_conclustions.tex
\section{Conclusions}
We have investigated the properties of recently proposed distributed word representations, which have shown results in several machine learning applications.  Despite their usefulness, understanding the mechanisms which afford them their characteristics is still a hard problem.

In this work, we presented an approach for viewing word embeddings as a language network.  We examined the characteristics of the induced networks, and their community structure.  Using this analysis, we were able to develop a procedure which develops a connected graph with meaningful clusters.  We believe that this work will set the stage for advances in both NLP techniques which utilize distributed word representations, and in understanding the properties of the machine learning processes which generate them.

Much remains to be done.  In the future we would like to focus on comparing word embeddings to other well known distributional representation techniques (e.g. LDA/LSA), examining the effects of different vocabulary types (e.g. topic words, entities) on the induced graphs, and the stability of the graph properties as a function of network size.

\section*{Acknowledgments}

This research was partially supported by NSF Grants DBI-1060572 and
IIS-1017181, with additional support from TASC Inc, and a Google Faculty Research Award.

%% file: 06_appendix.tex
\begin{figure}[p!]
        \centering
        \begin{subfigure}[b]{0.48\textwidth}
                \includegraphics[width=\textwidth]{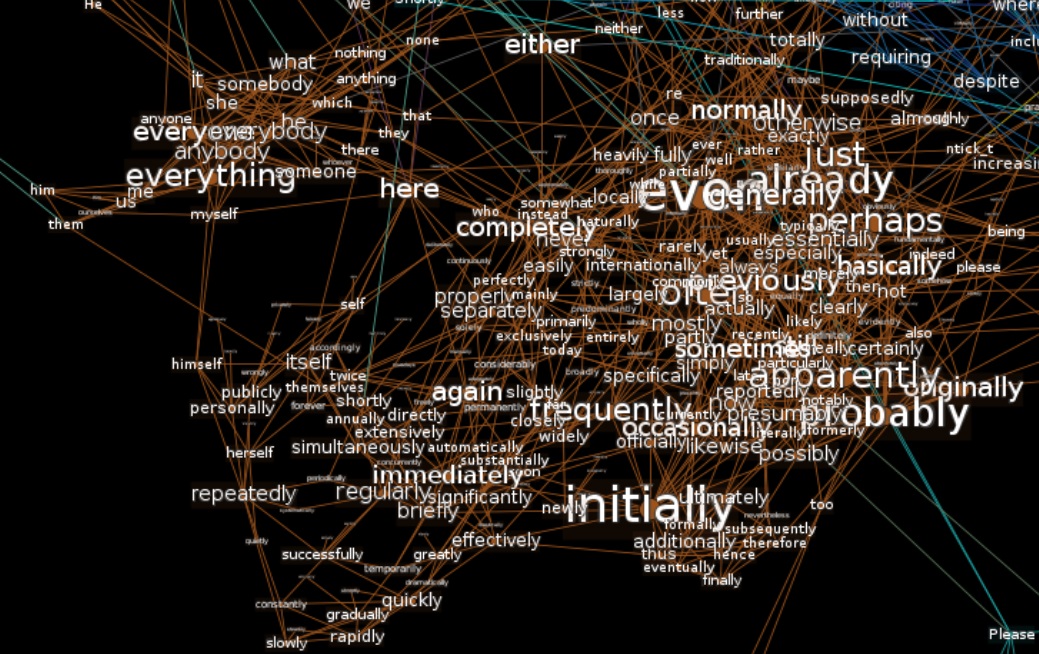}
                \caption{Pronouns \& Adverbs}
        \end{subfigure}
        ~
        \begin{subfigure}[b]{0.48\textwidth}
                \includegraphics[width=\textwidth]{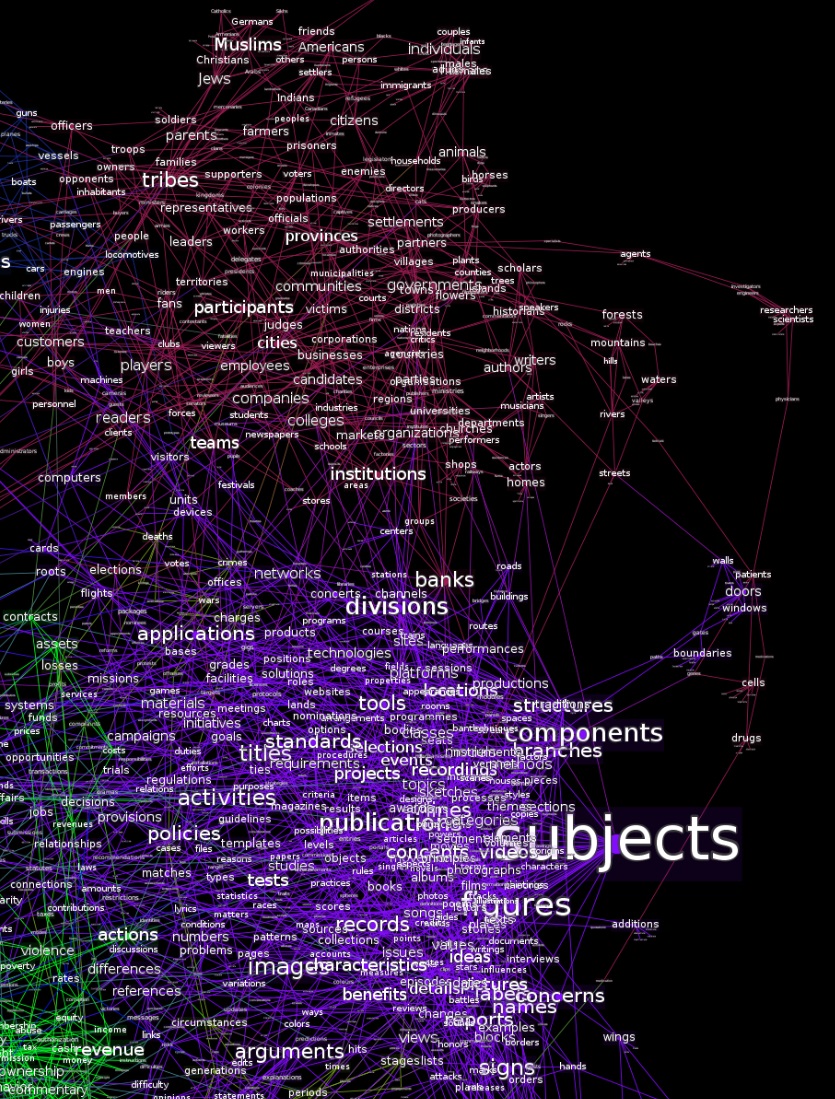}
                \caption{Plurals}
        \end{subfigure}                 
        \caption{Additional close-ups of clusters in Polyglot embeddings (from Figure \ref{fig:polyglot_en})}\label{fig:additional_polyglot_clusters}
\end{figure}

\begin{figure}[p!]
        \centering
        \includegraphics[width=\textwidth]{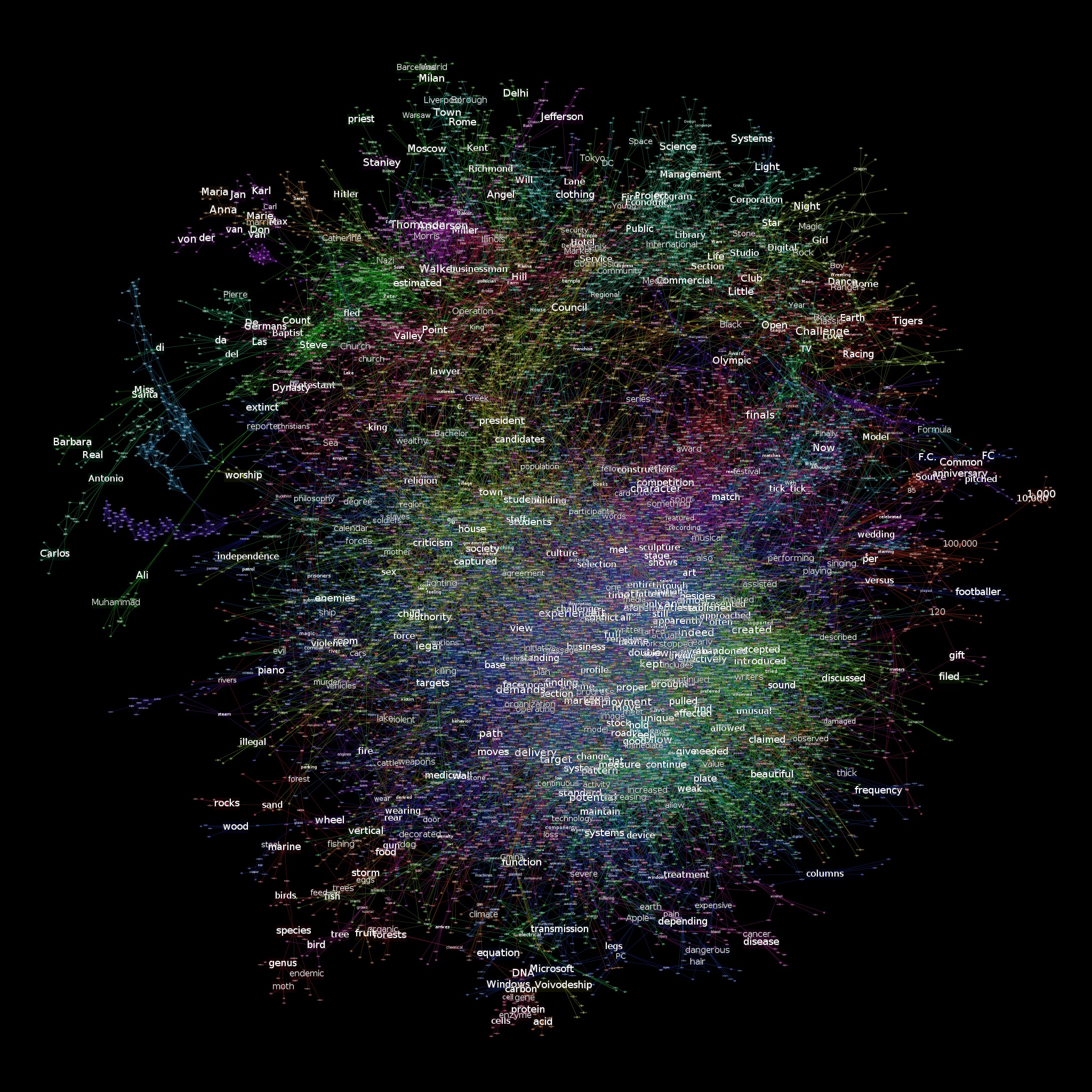}
        \caption{Visualization of the 6-NN  for the GCC of the top 5,000 most frequent words in the SkipGram embeddings.  SkipGram's representations are more semantic, and  so language features like polysemous words make global visualization harder.}\label{fig:skipgram_6nn}
\end{figure}